\documentclass{article} 
\usepackage{iclr2016_conference,times}
\usepackage{hyperref}
\usepackage{url}

\usepackage{graphicx}
\usepackage{threeparttable}
\usepackage{booktabs}
\usepackage{multirow}
\usepackage{tikz}
\usepackage{subfigure}
\usepackage{amsmath}
\usepackage{amssymb}

\usepackage[titletoc,title]{appendix}

\newcommand{\sn}[1]{\textsuperscript{#1}}

\title{Learning to Retrieve Out-of-Vocabulary Words in Speech Recognition}

\author{Imran Sheikh$^{123}$, Irina Illina$^{123}$ \& Dominique Fohr$^{123}$ \\
$^1$Universit\'{e} de Lorraine, LORIA, UMR 7503, Vandoeuvre-l{\`e}s-Nancy, F-54506, France \\
$^2$Inria, Villers-l{\`e}s-Nancy, F-54600, France	\\
$^3$CNRS, LORIA, UMR 7503, Vandoeuvre-l{\`e}s-Nancy, F-54506, France \\
\texttt{\{imran.sheikh, irina.illina, dominique.fohr\}@loria.fr} \\
\And
Georges Linar{\`e}s \\
Laboratoire Informatique d'Avignon \\
University of Avignon\\
\texttt{georges.linares@univ-avignon.fr} \\
}

\iclrfinalcopy 

\begin{document}

\maketitle

\begin{abstract}
Many Proper Names (PNs) are Out-Of-Vocabulary (OOV) words for speech recognition systems used to process diachronic audio data. To help recovery of the PNs missed by the system, relevant OOV PNs can be retrieved out of the many OOVs by exploiting semantic context of the spoken content. In this paper, we propose two neural network models targeted to retrieve OOV PNs relevant to an audio document: (a) Document level Continuous Bag of Words (D-CBOW), (b) Document level Continuous Bag of Weighted Words (D-CBOW2). Both these models take document words as input and learn with an objective to maximise the retrieval of co-occurring OOV PNs. With the D-CBOW2 model we propose a new approach in which the input embedding layer is augmented with a \emph{context anchor layer}. This layer learns to assign importance to input words and has the ability to capture (task specific) key-words in a bag-of-words neural network model. With experiments on French broadcast news videos we show that these two models outperform the baseline methods based on raw embeddings from LDA, Skip-gram and Paragraph Vectors. Combining the D-CBOW and D-CBOW2 models gives faster convergence during training.
\end{abstract}

\section{Introduction}
\label{sec:intro}
\emph{Large Vocabulary Continuous Speech Recognition} (LVCSR) based automatic audio indexing approaches allow search, navigation, browsing and structuring of large audio-video datasets based on spoken content \citep{google09,seide08}, as opposed to phonetic audio mining approaches which mostly serve user query based audio document retrieval from relatively smaller databases \citep{mindex}. 
However, LVCSR processing of diachronic audio data, and specifically broadcast audio news, can be challenging due to the variations in linguistic content and vocabulary. 
Thus leading to \emph{Out-Of-Vocabulary} (OOV) words for LVCSR. Even if good amount of training data is available, appending the LVCSR vocabulary and updating the \emph{Language Model} (LM) is not always a feasible solution \citep{LQinThesis}. 
Proper Names (PNs), which are important indexes for audio-video, have been found to be a major percentage of OOV words.
In this paper, we focus on the problem of retrieval of OOV PNs relevant to an audio document. 

To retrieve OOV PNs relevant to an audio document we rely on semantic context. In training phase, diachronic text news with new (i.e., OOV) PNs are collected from the internet. These set of text documents, referred as \emph{diachronic text corpus}, is used to learn a context vector space which captures relationship between the \emph{In-Vocabulary} (IV) words \& PNs and the OOV PNs. During test, the LVCSR hypothesis of the audio document is projected into the context space and then relevant OOV PNs are inferred. Recently it has been shown in \cite{icassp15} that \emph{Latent Dirichlet Allocation} (LDA) based topic space can perform with a good recall rate for retrieval of the \emph{target OOV PNs}\footnote{For a given audio documents several OOV PNs can be relevant, but only few of them are actually present in the audio document. We refer to the actual OOV PNs present in the audio document as \emph{target OOV PNs}}. In LDA, word and topic representations are constructed by counting word co-occurrences. Alternative methods to learn word and context representations \citep{NIPS2013_5021, pennington-socher-manning:2014:EMNLP2014}, based on predicting the context in which words appear, have become popular. These representations, also called \emph{embeddings}, have been shown to perform effectively when applied as pre-trained features in a range of applications and tasks \citep{baroni-dinu-kruszewski:2014:P14-1}.

In this paper, we propose two models targeted to retrieve OOV PNs relevant to an audio document: (a) Document level Continuous Bag of Words (D-CBOW) context model (b) Document level Continuous Bag of Weighted Words (D-CBOW2) context model. Both these models use pre-trained Skip-gram word embeddings \citep{NIPS2013_5021} for IV words \& PNs and learn with an objective to maximise the retrieval of the target OOV PNs in the document. The D-CBOW model gives equal importance to all the input terms (IV words \& PNs) co-occurring with the OOV PN in the document. In the D-CBOW2 model, we propose to augment the input embedding layer with a \emph{context anchor layer} which learns to assign importance to input words/features. This mechanism has the ability to capture (task specific) key-words/key-features in a bag-of-words neural network model.  With experiments on French broadcast news videos, we show that these two models outperform the baseline methods based on raw embeddings from LDA, Skip-gram and Paragraph Vectors \citep{icml2014c2_le14}. Additionally, D-CBOW2+ model, which combines the context vectors from the D-CBOW and D-CBOW2 models, gives faster convergence during training.

\subsection{Related Work}
OOV word recovery and vocabulary selection in LVCSR have traditionally relied on web search and selection methods based on frequency/recency of words \citep{icassp15}. The task of retrieval of OOV and PNs relevant to an audio document has been presented previously. 
\cite{senayICASSP13, senayIS13} use one LDA/LSA context model per PN which restricts the approach to  frequent PNs. \cite{icassp15,interspeech15a} presented methods based on probabilistic topic models and addressed the problems in ranking PNs arising due to bias in the probabilistic topic models. These methods readily apply to audio documents with single or coherent events. \cite{interspeech15b} applied word embeddings to audio documents with multiple events appearing one after another. As compared to these works we propose new models D-CBOW, D-CBOW2 and D-CBOW2+ trained to maximise the performance of retrieval of OOV PNs for audio documents with single/coherent events. 

Different extensions and variations in the \emph{Log-bilinear} (LBL) model and specifically the CBOW/Skip-gram architecture \citep{NIPS2013_5021} have been proposed for different tasks \citep{icml2014c2_le14,ling2015two,DBLP:journals/corr/NiuD15,DBLP:conf/aaai/QiuCNYR14,levy2014dependencybased,chen-liu-sun:2014:EMNLP2014}. For our task we propose document level bag-of-words architectures mainly because they are suitable to process LVCSR transcriptions of audio documents, which are firstly prone to noise due to word errors and secondly have no direct information about position of OOVs. Our novel contribution is the context anchor layer which learns to assign importance to input words and has the ability to capture (task specific) key-words in a bag-of-words neural network model. The context anchor layer is inspired by the attention mechanism presented in \cite{bahdanau2014neural}. Increasing importance of words in classical bag-of-words model of text has been discussed in \cite{impWords} for the task of text relatedness. The Deep Structured Semantic Model (DSSM) with convolutional-pooling structures presented in \cite{gao2014modeling} has been shown to capture keywords in text documents. 

\section{Methods to Represent and Retrieve OOV Proper Names}
\label{sec:methods}
As mentioned, our task is to retrieve OOV PNs relevant to an audio document. To achieve this we aim to learn a context vector space which captures relationship between the \emph{In-Vocabulary} (IV) words \& PNs and the OOV PNs, using a diachronic text corpus from the internet. During test, the LVCSR hypothesis of the audio document is projected into the context space to infer relevant OOV PNs. In this section we first briefly present the baseline method based on LDA, originally proposed in \cite{icassp15}. Then we present an extension of this method to raw Skip-gram and Paragraph Vector embeddings. Followed by the two proposed models, D-CBOW and D-CBOW2.

\subsection{Topic Space Representation based on LDA}
\label{sec:lda}
LDA topics are trained on the diachronic text corpus of ($D$) documents. Topic vocabulary size ($N$), the number of topics ($T$) and Dirichlet priors ($\alpha$, $\beta$) are first chosen. Topic model parameters $\theta$ and $\phi$ are then estimated using Gibbs sampling algorithm \citep{Griffiths06042004}. $\theta = [\theta_{dt}]_{D \times T}$ is the topic distribution for each document $d$, and $\phi = [\phi_{vt}]_{{N} \times T}$ is the topic distribution to words from the vocabulary, both across $T$ topics. 
Let us denote the LVCSR word hypothesis by $h$ and OOV PNs in diachronic corpus (and topic model vocabulary) by $\tilde{v}_i$. The latent topic mixture of $h$, i.e. $p(t | h)$, is inferred by sampling the topic assignments for words in $h$ using the word-topic distribution $\phi$ learned during training. Given $p(\tilde{v}_i | t)$ from $\phi$, the likelihood of an OOV PN ($\tilde{v}_i$) in the diachronic corpus is calculated as:
\begin{equation}
\label{eq:rt1}
p(\tilde{v}_i | h) = \sum_{t=1}^{T} \; p(\tilde{v_i} | t) \;p(t | h)
\end{equation} 
To retrieve OOV PNs we calculate $p(\tilde{v}_i | h)$ for each $\tilde{v}_i$ and then use it as a score to rank OOV PNs relevant to $h$. 

\subsection{Raw Word Embeddings from Paragraph Vector and Skip-gram Models}
\label{sec:w2v}
\cite{icml2014c2_le14} proposed Paragraph Vector - a distributed model to represent sentences, paragraphs and documents. For our task, this model can represent test documents and OOV PNs in a common vector space and we study its performance to retrieve relevant OOV PNs.
The $K$ dimensional vector representation of the LVCSR hypothesis of the audio document ($h$) is compared with the embeddings ($\tilde{v}_i$) for each of the OOV PNs to calculate a score as follows:
\begin{equation}
\label{eq:rt2}
s \approx \max_{i} \{ CosSim(h, \tilde{v}_i) \}  
\end{equation}
where $CosSim(\cdot, \cdot)$ is the cosine similarity measure. The score $s$ is used to rank and retrieve OOV PNs relevant to test document. 

We also examine a simple alternative to Paragraph Vectors. In this method, during training Skip-gram word embeddings\footnote{Word embeddings from Skip-gram model give better performance than those from the CBOW model proposed in the original work \cite{NIPS2013_5021}.} are learned for all the words in the diachronic corpus. Given these word embeddings and their linearity property, we obtain a representation for a test document by taking an average over all the vocabulary words in the document. This document representation is referred to as \emph{AverageVec}. With the AverageVec test document representation ($h$) and the Skip-gram embeddings of the OOV PNs ($\tilde{v}_i$), the OOV PNs relevant to the test documents can be retrieved using Equation \ref{eq:rt2}.

\subsection{Learning OOV Context to Maximise Retrieval Performance}
\label{sec:dbow}
We propose two models targeted to retrieve OOV PNs relevant to an audio document. These models operate at document level and learn with an objective to maximise the retrieval of the target OOV PNs in the document. We choose document level input as it is suitable to process LVCSR transcriptions of audio documents, which are prone to noise due to word errors and have no direct information about position of OOVs.

\subsubsection{Document level Continuous Bag of Words (D-CBOW) Context Model}
\label{sec:dcbow}
\begin{figure}[h]
\begin{center}
\includegraphics[width=0.8\linewidth]{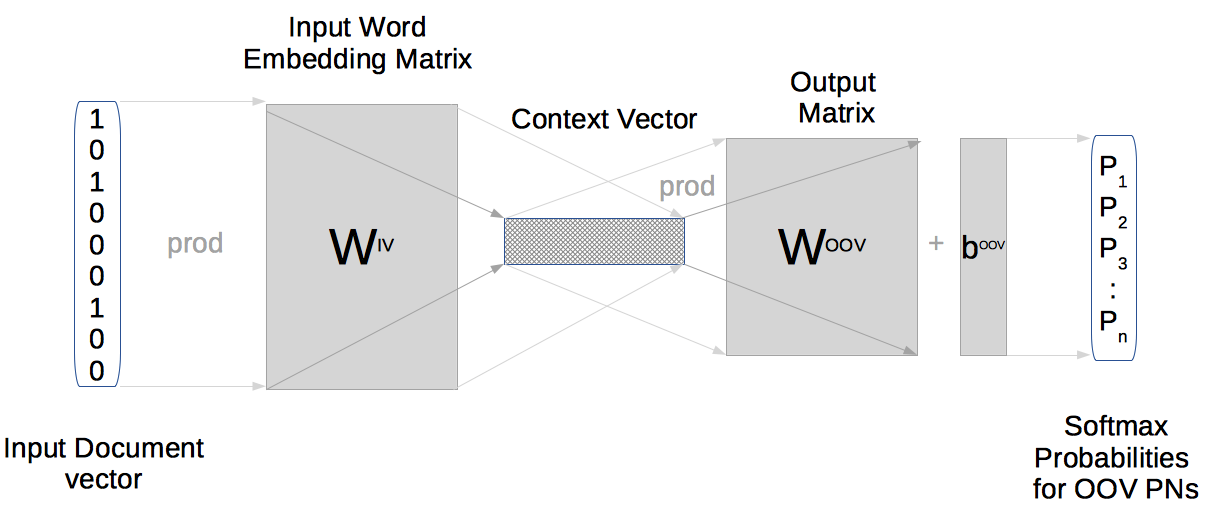}
\end{center}
\caption{Document level Continuous Bag of Words (D-CBOW) Model}
\label{fig:dcbow}
\end{figure}

Figure \ref{fig:dcbow} depicts a forward pass of the proposed D-CBOW model. The D-CBOW model takes at input all the IV words \& PNs in a document. The context vector ($c_d$) for an input document ($d$) is obtained using the matrix-vector product between the input word embedding matrix $W^{IV}$ and the input document vector. During training, the co-occurring OOV PNs in the document are set at the output. (The cost function used to train the network shares similarity with that of CBOW model originally proposed in \cite{NIPS2013_5021}, we refer the interested readers to \cite{DBLP:journals/corr/Rong14}.) During test, the softmax probabilities at the output are used as scores to rank and retrieve the OOV PNs. 

To train the D-CBOW model we first learn Skip-gram embeddings for the IV words \& PNs. The $W^{IV}$ parameter in the model is set with these embeddings. In the first training phase, only the 
$W^{OOV}, b^{OOV}$ parameters are trained 
(until convergence on the validation set), keeping the $W^{IV}$ parameter fixed. After this first training phase, all the 
model parameters ($W^{IV}, W^{OOV}, b^{OOV}$) 
are trained and updated in the second training phase. In our experiments we observed that the model trained with all the network parameters learned in a single training phase gives lower retrieval performance compared to the model trained in two phase training. 

\subsubsection{Document level Continuous Bag of Weighted Words (D-CBOW2) Context Model}
\label{sec:dcbow2}
\begin{figure}[h]
\begin{center}
\includegraphics[width=0.8\linewidth]{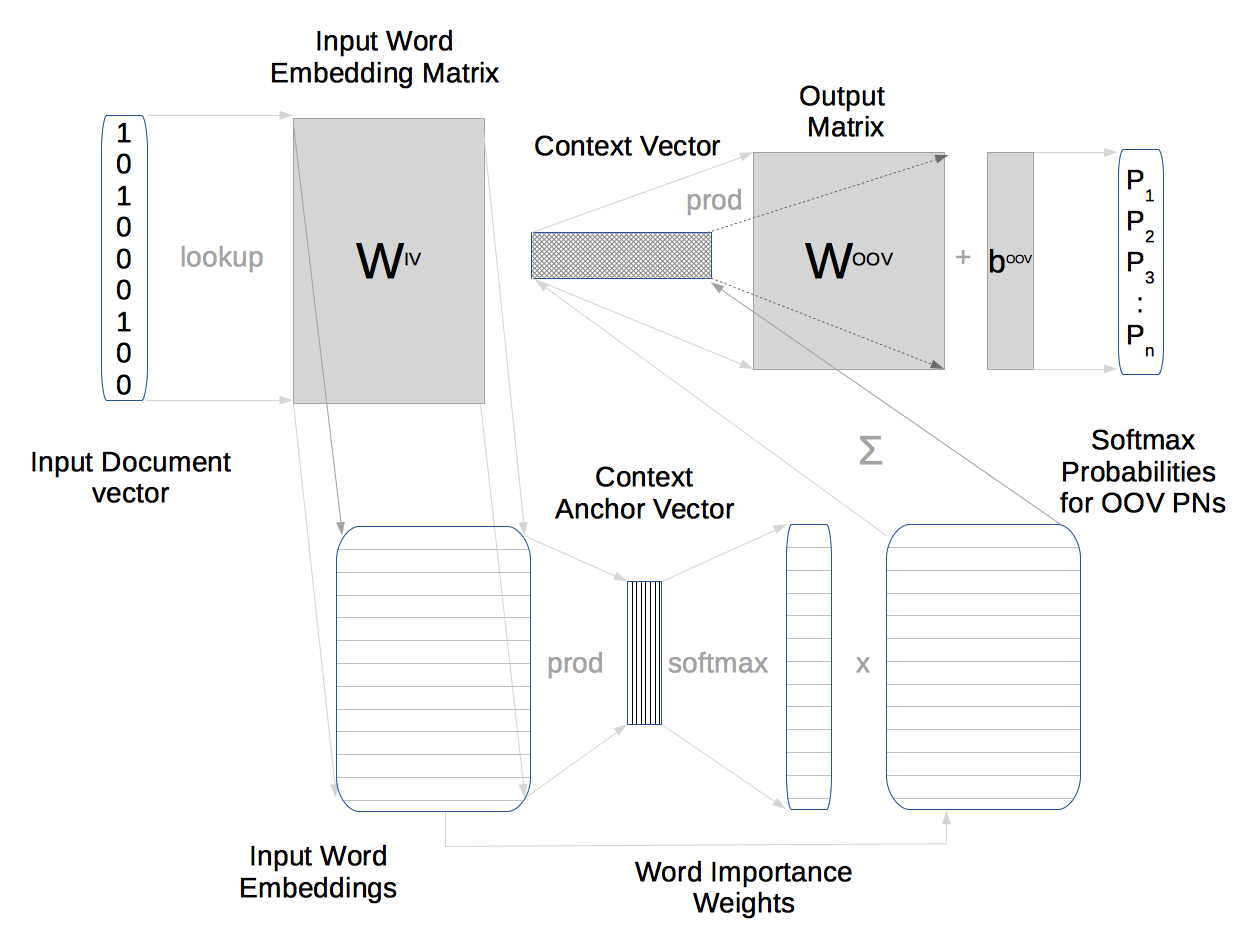}
\end{center}
\caption{Document level Continuous Bag of Weighted Words (D-CBOW2) Model}
\label{fig:dcbow2}
\end{figure}

Figure \ref{fig:dcbow2} depicts a forward pass of the proposed D-CBOW2 model. Like the D-CBOW model, the D-CBOW2 model also takes at input all the IV words and PNs in a document. During training the co-occurring OOV PNs in the document are set at the output and during test the softmax probabilities at the output are used as scores to rank the OOV PNs. While the D-CBOW model gives equal importance to the input embeddings, the D-CBOW2 model learns to assign importance to input embeddings. To achieve this it has a context anchor vector which itself is learned during training. The importance weight ($\omega_j$) for the $j^{th}$ input embedding ($w^{IV}_{j}$) is calculated as:
\begin{equation}
\label{eq:wts}
\begin{split}
\delta_j &= prod(w^{IV}_{j}, C^{A}) \\
\omega_j &= \frac{exp(\delta_j)}{\sum_{j} exp(\delta_j)}
\end{split}
\end{equation}
where $C^{A}$ denotes the context anchor and $prod(\cdot, \cdot)$ is a matrix product (a matrix-vector product in this case). Given the importance weights, the context vector ($c_d$) for an input document ($d$) is calculated as a weighted sum of the input embeddings as follows:
\begin{equation}
\label{eq:wsum}
c_d = \sum^{N_d}_{j} \omega_{j} \; w^{IV}_{j}
\end{equation}
The steps to train the network are similar to D-CBOW. The $W^{IV}$ parameter in the model is set with Skip-gram embeddings. In the first training phase, the 
$W^{OOV}, b^{OOV}, C^{A}$ 
parameters are trained, keeping the $W^{IV}$ parameter fixed. After this first training phase, all the model parameters 
($W^{IV}, W^{OOV}, b^{OOV}, C^{A}$) 
are trained and updated in the second training phase. 

\section{Experiments and Results}
\label{sec:exp}
\subsection{Experiment Corpus}
Table \ref{tab:data} presents realistic broadcast news diachronic datasets which will be used as the train, validation and test sets in our study. These datasets also highlight the motivation for handling OOV PNs. The datasets are collected from two different sources: (a) French newspaper \emph{L'Express} (http://www.lexpress.fr/) (b) French website of the \emph{Euronews} (http://fr.euronews.com/) television channel. The \emph{L'Express} dataset contains text news whereas the Euronews dataset contains text news as well news videos and their text transcriptions. In our study the \emph{L'Express} dataset will be used as diachronic corpus to train context/topic models, in order to infer the OOV PNs relevant to Euronews videos. \emph{Euronews} text documents, denoted as 'validation' in Table \ref{tab:data}, are used as a validation set in our experiments.

TreeTagger \citep{TreeTagger} is used to automatically tag PNs in the text. The words and PNs which occur in the lexicon of our \emph{Automatic News Transcription System} (ANTS)\citep{ANTS2004} are tagged as IV and remaining PNs are tagged as OOV. ANTS lexicon is based on news articles until 2008 from French newspaper \emph{LeMonde} and contains about 123K unique words. As shown in the Table \ref{tab:data}, 72\% of OOV words in Euronews video dataset are PNs and about 64\% of the videos contain OOV PNs. The total number of OOV PNs to be retrieved for the \emph{Euronews} videos, obtained by counting unique OOV PNs per video, is 4694. Out of 4694, up to 2010 (42\%) OOV PNs can be retrieved with the training (\emph{L'Express}) diachronic corpus. As shown in the Table~\ref{tab:data} the training corpus has 9.3K new (OOV) PNs to learn. This number and the target OOV PN coverage can be increased by augmenting text documents from additional sources \citep{lrec16}, for instance from other news websites. However this is not the focus of our study in this paper. 

\begin{table*}[h!tb]
  \centering
  \begin{threeparttable}[b]
  \caption{Broadcast news diachronic datasets}
  \label{tab:data}
  \begin{tabular}{ l c  c  c  c   }
 \toprule
 \multirow{2}{*}{}				& \emph{L'Express} 	& \emph{Euronews}  & \emph{Euronews}	\\ 
 							& 	(train)	       	&     (validation)     	&  (test)	        		\\ 							
 \midrule
 Type of Documents  			& Text			& Text 	          	& Video 			\\ 
  Time Period 	        				& Jan - Jun 2014	& Jan - Jun 2014  	& Jan - Jun 2014	\\
 Number of Documents\sn{1}	 	& 45K 			& 3.1K 	  		& 3K				\\ 
 Vocabulary Size (unigrams)\sn{2} 	& 150K 			& 42K		  	& 45K			\\ 
 Corpus Size (approx. word count) 	& 24M 			& 550K	  		& 700K 			\\ 
 \midrule
 Number of PN unigrams\sn{2} 		& 57K 			& 12K		  	& 11K			\\ 
 Total PN count 				& 1.45M 			& 54K		  	& 42K			\\ 
  \midrule
 Number of OOV unigrams\sn{3} 	& 12.4K 			& 4.9K		  	& 4.3K			\\ 
 Documents with OOV\sn{3} 		& 32.3K 			& 2.25K	  		&2.2K			\\ 		
 Total OOV count\sn{3} 			& 141K 			& 9.1K		  	& 8K				\\ 		
 \midrule
 Number of OOV PN unigrams\sn{3} & 9.3K 			& 3.4K		  	& 3.1K			\\   
 Documents with OOV PN\sn{3} 	& 26.5K 			& 1.9K		  	& 1.9K			\\ 	 
 Total OOV PN count\sn{3} 		& 107K 			& 6.9K		  	& 6.2K			\\ 	
 \bottomrule		 
  \end{tabular}
\begin{tablenotes}
\item \sn{1}K denotes \emph{Thousand} and M denotes \emph{Million} \\
\sn{2} In \emph{L'Express} unigrams occurring less than two times are ignored \\
\sn{3} In \emph{L'Express} unigrams occurring in less than three documents ignored; documents with less than 20 and more than 500 terms are removed\\
Note: OOV and OOV PN statistics are post term-document filtering
    \end{tablenotes}
  \end{threeparttable}
\end{table*}

\subsection{Experiment Setup}
The ANTS \citep{ANTS2004} LVCSR system is used to perform automatic segmentation and speech-to-text transcription of the test set (\emph{Euronews}) audio news. The automatic transcriptions of the test audio news obtained by ANTS have an average \emph{Word Error Rate} (WER) of 40\% as compared to the reference transcriptions available from \emph{Euronews}. 

For training the context/topic models, diachronic corpus words are lemmatised and filtered by removing PNs and non PN words occurring less than 3 times. Additionally a stop-list of common French words and non content words is used. Moreover, a POS based filter is employed to choose words tagged as PN, noun, adjective, verb and acronym. PNs not present in the ANTS LVCSR lexicon and are tagged as OOV PNs. The context and topic models are trained with this filtered vocabulary. For comparison, the number of LDA topics and the dimensionality of the different neural context models are chosen to be equal and set to 100. A window size of 15 is chosen for training the Skip-gram embeddings. This selection is based on performance on the validation set.

Our baseline methods, discussed in Section \ref{sec:lda} and Section \ref{sec:w2v}, are denoted as LDA, Doc2Vec (for Paragraph Vector method), AverageVec. Our proposed models, discussed in Section \ref{sec:dcbow} and Section \ref{sec:dcbow2}, are denoted as D-CBOW and D-CBOW2. Additionally we present the results of our experiments with a combination of the D-CBOW and D-CBOW2 models. We denote it as D-CBOW2+. In D-CBOW2+ model the bag-of-words and the bag-of-weighted-words context vectors are concatenated together. 

\subsection{Training the D-CBOW group of models}
\begin{figure*}[ht]
\begin{minipage}[b]{1\linewidth}
  \centering
 \includegraphics[width=0.5\textwidth]{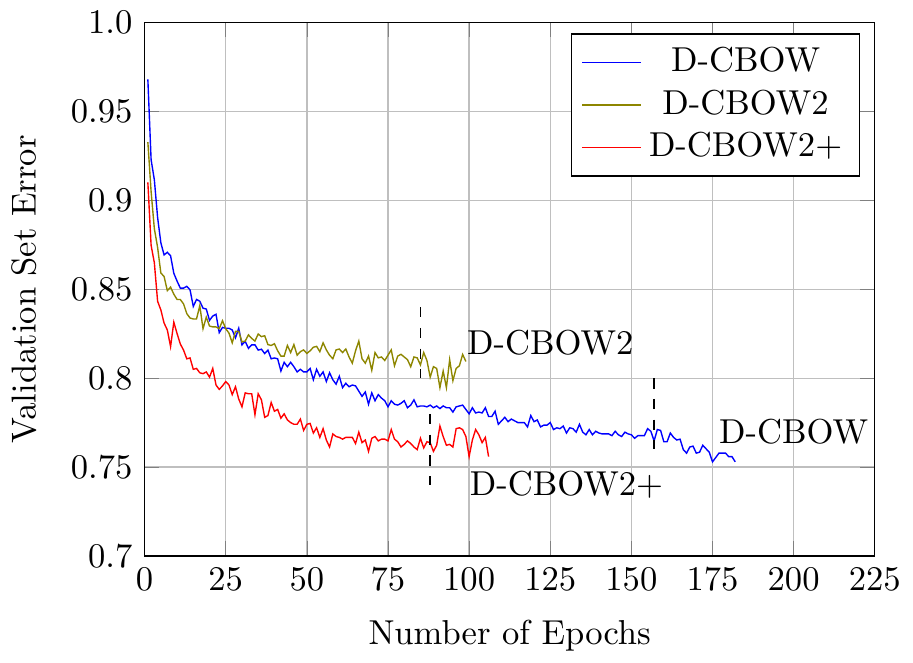}
\end{minipage}
\caption{Validation set error during training of D-CBOW, D-CBOW2 and D-CBOW2+ models. ($\mbox{- - -}$ markers indicate end of first training phase and begin of second training phase)}
\label{fig:err}
\end{figure*}

As discussed in Section \ref{sec:dcbow} and Section \ref{sec:dcbow2}, the D-CBOW and D-CBOW2 models are trained in two phases. The D-CBOW2+ model is also trained in two phases. To control the training of the D-CBOW, D-CBOW2 and D-CBOW2+ models an early stopping criterion \citep{DBLP:journals/corr/abs-1206-5533} based on the validation set error is used. Early stopping is applied in both the first and the second training phases. The patience limits in early stopping are different for the two training phases. Figure \ref{fig:err} shows a graph of validation set errors, of the D-CBOW, D-CBOW2 and D-CBOW2+ models, as the training progresses. It must be noted that this error is like a classification error and a measure of whether an OOV PN in the validation set document is given the highest output probability or not. The actual performance measures used to evaluate our OOV PN retrieval task are different and discussed in the next section. The classification error is shown as it is easier to analyse and visualise the training of the models. For instance it can be observed in Figure \ref{fig:err} that the D-CBOW2+ model gives a faster convergence compared to the D-CBOW and D-CBOW2 models. We analyse the retrieval performance and the training convergence of these models in detail in Section \ref{sec:cmp}.

\begin{figure*}[ht]
\begin{minipage}[b]{1\linewidth}
  \centering
 \includegraphics[width=\textwidth]{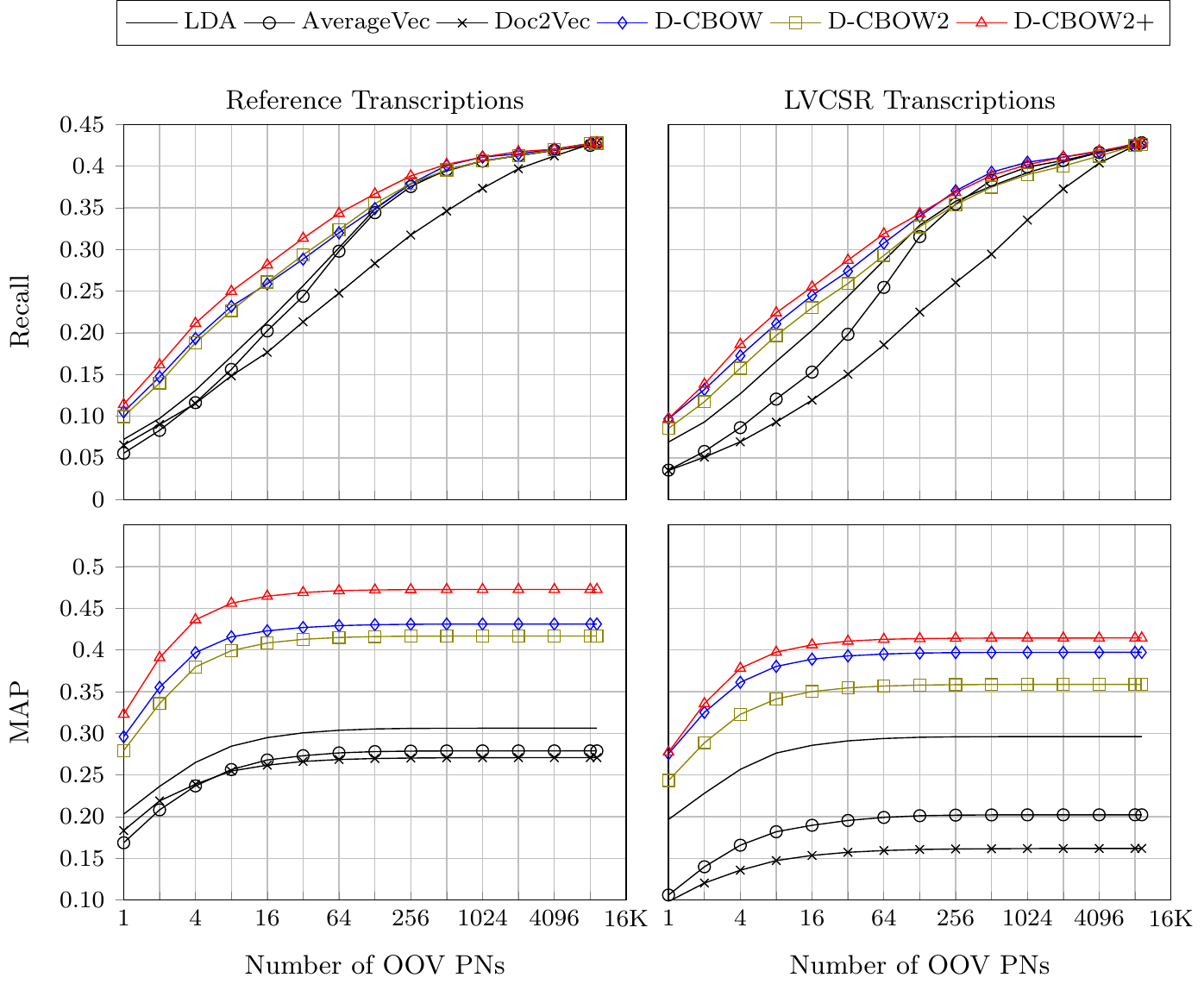}
\end{minipage}
\caption{Recall (top) \& MAP (bottom) performance of OOV PN retrieval for the reference (left) and LVCSR (right) transcriptions of the \emph{Euronews} video test set. (For D-CBOW, D-CBOW2 and D-CBOW2+ the evaluation is done after the first training phase.)}
\label{fig:rec2}
\end{figure*}

\begin{figure*}[ht]
\begin{minipage}[b]{1\linewidth}
  \centering
 \includegraphics[width=\textwidth]{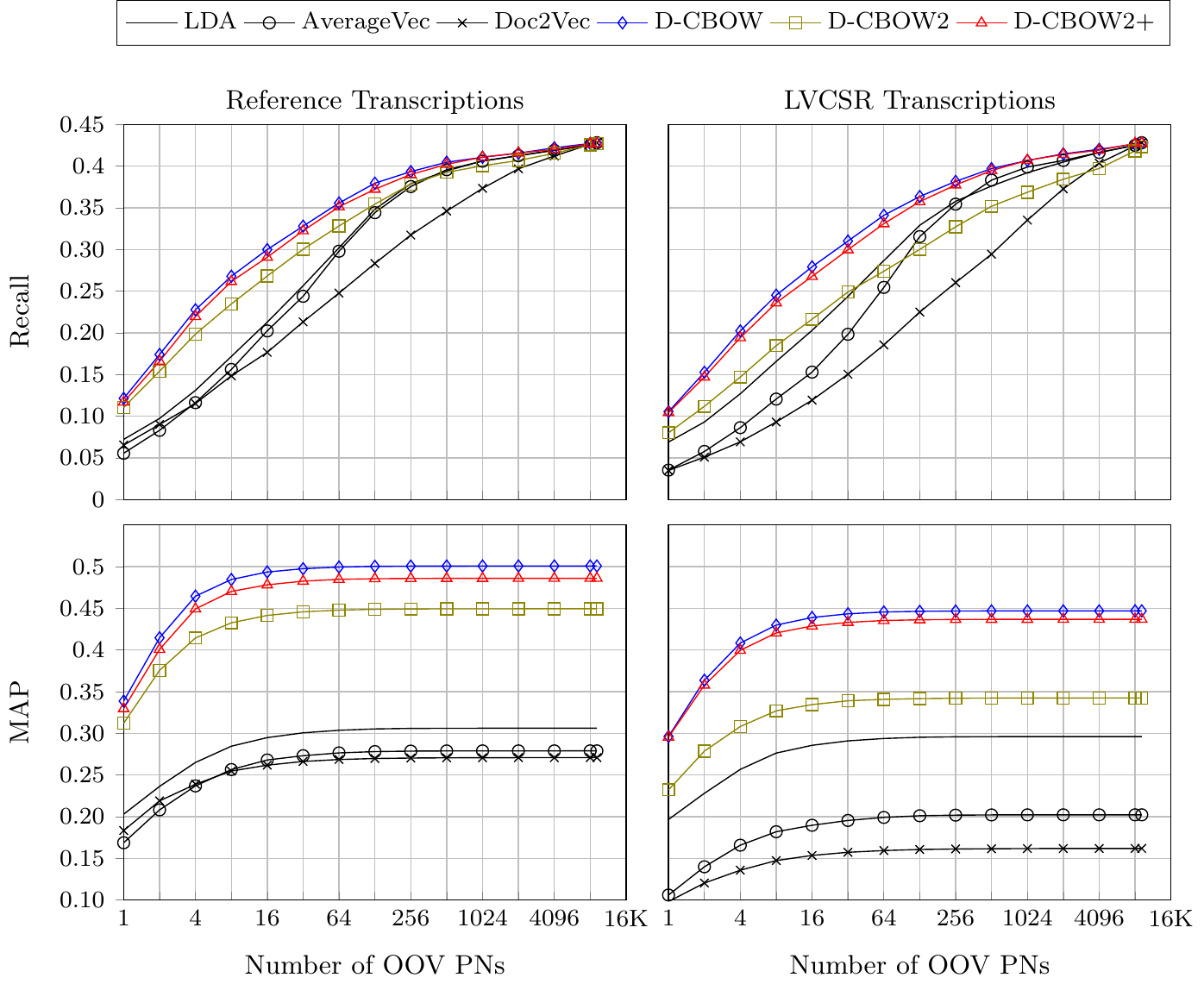}
\end{minipage}
\caption{Recall (top) \& MAP (bottom) performance of OOV PN retrieval for the reference (left) and LVCSR (right) transcriptions of the \emph{Euronews} video test set. (For D-CBOW, D-CBOW2 and D-CBOW2+ the evaluation is done after the second training phase.)}
\label{fig:rec}
\end{figure*}

\subsection{OOV PN Retrieval Performance}
Figure \ref{fig:rec2} and Figure \ref{fig:rec} show the \emph{Recall} and \emph{Mean Average Precision} (MAP) \citep{ManningRaghavanSchuetze08} performance of retrieval of OOV PNs, for models discussed in Section \ref{sec:methods}. Figure \ref{fig:rec2} depicts the performance of D-CBOW, D-CBOW2 and D-CBOW2+ after the end of first training phase, whereas Figure \ref{fig:rec} depicts the performance after the end of second training phase. The graphs shown are for the reference transcriptions (left) and the LVCSR transcriptions (right) of the \emph{Euronews} test set audio. The X-axis represents the number of OOV PNs selected from the diachronic corpus i.e. the 'N' in the top-N retrieved results. Y-axis represents recall (top) and MAP (bottom) of the target OOV PNs. 

\section{Discussion}
\subsection{A note on Recall and MAP}
It is necessary to understand the importance and differences for the recall and MAP curves. After retrieval of the relevant OOV PNs, the top-N relevant OOV PNs are to be used for recovery/recognition of the target OOV PNs. To recover the target OOV PNs one can use phone matching \citep{Cheng05}, or additional speech recognition pass \citep{interspeech15b,Oger08}); or spotting PNs in speech \citep{Parada2010,kaldiKws}. In each of these approaches, the retrieval ranks/scores may or may not be used. This is where the recall and MAP curves make a difference. The recall value at an \emph{operating point} (N in the top-N choice) is not sensitive to the rank of the retrieved OOV PNs whereas the MAP value, by definition, takes into account the retrieval ranks. For instance if we take top 5\% (top 465 retrieved OOV PNs\footnote{5\% of 9.3K. This might appear as a small number but it must be noted that as we increase the diachronic corpora to increase coverage of target OOV PNs this number will also multiply \citep{lrec16}.}) of the retrieved OOV PNs almost all the methods will have same recall, but the MAP will have  differences. In Section \ref{sec:kws} we try to analyse the effect of the recall and MAP of the retrieval task on recovery of the target OOV PNs. To recover the target OOV PNs in the audio documents we perform a keyword search on the LVCSR lattices using the approach proposed by \citet{kaldiKws}. This is a quick and simple approach to evaluate the retrieved list of top-N OOV PNs.

\subsection{Comparison of Retrieval Performance}
\label{sec:cmp}
Among the baseline methods, LDA based method performs better than the methods based on AverageVec and Doc2Vec embeddings\footnote{It is also interesting to note that the LDA based method shows least degradation due to LVCSR errors.}. AverageVec performs better than Doc2Vec, specially for LVCSR transcriptions. The proposed D-CBOW and D-CBOW2+ models clearly outperform the baseline methods, both for reference and LVCSR transcriptions. Performance of D-CBOW2+ is similar to that of D-CBOW, but as mentioned earlier D-CBOW2+ has faster convergence during training. The D-CBOW2 model also outperforms the baseline methods for reference transcriptions. For LVCSR transcriptions its performance is affected due to LVCSR errors in important keywords. We discuss more about this in Section \ref{sec:impwts}. 

Here we will try to discuss why and when D-CBOW2 and D-CBOW2+ models do not outperform the D-CBOW model. First we can observe from Figure \ref{fig:err} that the second training phase helps only the D-CBOW model. For D-CBOW2 model it causes the validation error to increase after a few epochs and for the D-CBOW2+ model the validation error is kind of oscillating; and thus the second training phase makes a quick early stop for both these models. We believe that this problem of convergence during the second training phase of D-CBOW2 and D-CBOW2+ models is because both the input embeddings and the context anchor vector (applied to the input embeddings to obtain importance weights) are being updated during the subsequent training min-batches. We believe that this takes the D-CBOW2 and D-CBOW2+ models away from convergence. Currently we are working on addressing this problem. Note that during the first training phase of D-CBOW2 and D-CBOW2+, only the context anchor vector is trained/updated at the input side. We find that D-CBOW2 model stops at a higher validation error even in the first training phase as it is constrained to classify OOV PNs based on handful of words (few words assigned high importance weights, remaining close to zero). D-CBOW2+ model on the other hand uses average of input words as well as the specific words assigned higher importance. This combination at the input side gives a faster convergence in the first training phase (majority of the training time), even though the model parameters (output layer) are doubled in size.

We can see the effect of this even on the retrieval performance. We have shown the Recall and MAP performance at the end of first training phase in Figure \ref{fig:rec2}. For better clarity we have also shown the difference in the Recall and MAP performances between the second and the first training phases in Figure \ref{fig:diff} in Appendix \ref{app:train1}. Firstly, it is interesting to note from Figure \ref{fig:rec2} that at the end of first training phase the D-CBOW2+ performs better than the D-CBOW model, and the D-CBOW2 model is not far from D-CBOW as after second training phase. Secondly, even the differences between performance of the LVCSR and reference transcriptions, for proposed models, is relatively less (about 0.5 to 0.6 MAP) as compared to that at the end of second training phase (about 0.5 to 1.0 MAP). Further from the Figure \ref{fig:diff} we can observe that the second training phase improves the Recall and MAP for D-CBOW and this improvement is more for the reference transcriptions. For D-CBOW2 improvements are seen in the reference transcriptions but these come at the cost of degradation in the Recall and MAP for LVCSR transcriptions. In case of D-CBOW2+, there are only small improvements for (both) reference and LVCSR transcriptions, again suggesting that the second training phase is not working with importance based models. 

\subsection{Keyword Importance by D-CBOW2 model}
\label{sec:impwts}
In Table \ref{tab:kw}, we demonstrate the ability of the D-CBOW2 model to give importance to keywords in the input document in order to infer the target OOV PN. We have listed four \emph{Euronews} video examples. The first column lists the target OOV PNs present in these videos. The words given importance by the D-CBOW2 models (importance weight more than 0.1 as calculated by Equation \ref{eq:wts}), from corresponding input reference and manual transcriptions are shown in second and fifth column respectively. For comparison the ranks given by the D-CBOW2 and D-CBOW models are also shown. It can be seen that in the first two examples the D-CBOW2 model extracts relevant keywords and gives similar ranks as D-CBOW model. In the third example, the ranks given by D-CBOW2 models are better than that of D-CBOW as the context, influenced by the keywords, is more focused. But relying on few words can cause degradation when (a) the LVCSR transcription loses some of the interesting words and (b) the target OOV PNs cannot be discriminated with the available interesting words. This can be seen in the fourth example, where words \emph{kiev} and \emph{parti} were mis-recognised during LVCSR transcription. And even though the word \emph{cfa}, in the LVCSR transcription, is a good descriptor of the topic of the input document it does lead to good ranks as it is not close to the target OOV PNs in the learned space. The DCBOW model on the other hand takes a decision based on average of the words in the LVCSR transcription and thus gets relatively better ranks.

\begin{table*}[h!tb]
  \centering
  \begin{threeparttable}[b]
  \caption{Keyword Importance and Ranks by D-CBOW2 model}
  \label{tab:kw}
  \begin{tabular}{ l c  c  c  c  c  c }
 \toprule
 Target						&  \multicolumn{3}{c}{Reference Transcription} 	&  \multicolumn{3}{c}{LVCSR Transcription}	\\ 
 \cmidrule(r){2-7}
OOV  PNs						& 	D-CBOW2	& \multicolumn{2}{c}{ranks} 	&  D-CBOW2	& \multicolumn{2}{c}{ranks}   		\\ 							
 							& 	keywords\sn{*}	 & R2\sn{\#}        	& R\sn{\#}  	&  keywords\sn{\#}	& R2\sn{\#}    & R\sn{\#}	\\ 							
 \midrule
  \multirow{3}{*}{\emph{kehm}}  			& \emph{ski chu coma}	 			& \multirow{3}{*}{1}  & \multirow{3}{*}{1} 	         & \emph{ski coma} 			& \multirow{3}{*}{1}  	& \multirow{3}{*}{1}		\\ 
  							& \emph{m{\'e}ribel}      			&			       &  					& \emph{m{\'e}ribel} 			& 				& 					\\	
  							& \emph{schumacher} 		     		&				&					&  					&				&					\\
 \midrule
\emph{mh370} 						& \emph{malaisienne}   			& 1  				& 2					& \emph{airlines} 		 		& 1  				& 1					\\ 
\emph{bluefin-21}						& \emph{d{\'e}bris}				& 3				& 3					& \emph{batterie}				& 5				& 3 					\\
							& \emph{radar}					& 				& 					& 					& 				&  					\\
\midrule
\emph{snowden} 						& \emph{d{\'e}put{\'e}}    			& 1 				& 1					& \emph{confidentialit{\'e}} 		& 5  				& 1					\\
\emph{roussef}						& \emph{dilma}					& 30 				& 5181 				& \emph{{\'e}douard}			& 718			& 1882				\\
  							& \emph{president} 		     		&				&					&  					&				&					\\
\midrule
\emph{zhukovskaya}					&\emph{kiev}  					& 286 			& 205				& \emph{cfa} 				& 5388 			& 333				\\
\emph{ma\"{i}dan}						& \emph{parti}					& 1				& 4 					& 					& 4780 			& 16 					\\
\emph{olesya} 						&						& 265 			& 211 				& 					& 5179 			& 346				\\
 \bottomrule		 
  \end{tabular}
\begin{tablenotes}
\item \sn{*}Input words getting importance weight more than 0.1 shown as keywords \\
\sn{\#} R, R2 denote ranks given by D-CBOW, D-CBOW2 models 
\end{tablenotes}
  \end{threeparttable}
\end{table*}

\subsection{Recovery of Target OOV PNs with Automatic Keyword Search (KWS)}
\label{sec:kws}
\begin{figure*}[ht]
\begin{minipage}[b]{1\linewidth}
  \centering
 \includegraphics[width=0.99\textwidth]{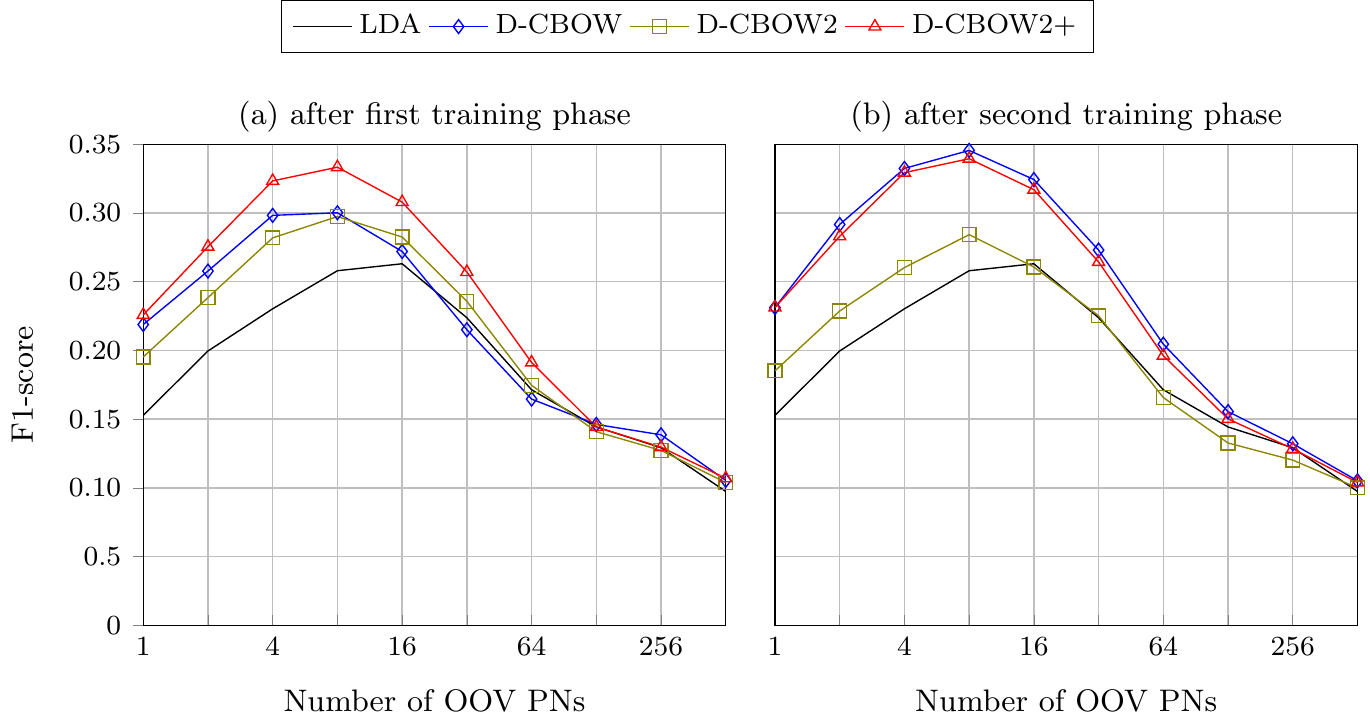}
\end{minipage}
\caption{$F_1$ scores for recovery of target OOV PNs with different representations.}
\label{fig:f1}
\end{figure*}

The recovery of the target OOV PNs in a diachronic audio is done in two steps. First, a list of relevant OOV PNs is retrieved with the models presented in this paper. In the second step, an automatic Keyword Search (KWS) is performed, for each OOV PN in the list of relevant OOV PNs, on the entire LVCSR lattice of the audio file. For this KWS we employ the approach proposed by \citet{kaldiKws}, which enables searching of OOV words (in their phonetic form) in an LVCSR lattice. The OOV PN recovery performance is evaluated in terms of \emph{$F_1$-score}, calculated as:
\begin{align}
\begin{split}
\label{eq:f1}
F_{1} &= \frac{2*precision*recall}{precision+recall} \\
       &= \frac{2* tp}{2 * tp + fp + fn}
\end{split}
\end{align}
where $tp, fp, fn$ stand for number of true positive, false positive and false negative OOV PNs respectively. (Note that the precision and recall here correspond to the phonetic search and are different from those of the retrieval task discussed previously.) 

Figure \ref{fig:f1} shows the F1-scores for recovery of the target OOV PNs. For comparison of performance of D-CBOW, D-CBOW2 and D-CBOW2+ models, the F1-scores obtained with the OOV PN lists after the first and the second training phase are shown. Among the baseline models, F1-scores for only LDA based approach is shown because LDA performs the best. The X-axis represents the number of top-N relevant OOV PNs selected for keyword search. The keyword search algorithm has a matching score threshold which controls the operating characteristics (and hence recall/precision and F1-score) of the phonetic search. We show the best F1-scores corresponding to top-N OOV PN lists of different sizes (N). The F1-scores are shown only until top-512 OOV PNs because beyond this there is no significant difference in the F1-scores of the different retrieval methods. Overall we can observe that, better the Recall and MAP of OOV PN retrieval the better is the F-1 score. F1-scores are best for the list of OOV PNs retrieved by the proposed models D-CBOW and D-CBOW2+. They obtain the best F1-score of about 0.34 with the top-8 retrieved OOV PNs. At this F1-score the recall and precision for OOV recovery with D-CBOW is 0.30 and 0.38 respectively. Similarly for D-CBOW2+ it is 0.29 and 0.42 respectively



\subsubsection*{Acknowledgments}
This work is funded by the ContNomina project supported by the French National Research Agency (ANR) under contract ANR-12-BS02-0009.
The KWS experiments presented in this paper were carried out using the Grid'5000 testbed, supported by a scientific interest group hosted by Inria and including CNRS, RENATER and several Universities as well as other organizations (see https://www.grid5000.fr).

\bibliography{iclr2016_conference}
\bibliographystyle{iclr2016_conference}

\clearpage

\begin{appendices}

\section{Difference in OOV PN Retrieval Performance obtained after the First and the Second Training Phase}
\label{app:train1}
\begin{figure*}[ht]
\begin{minipage}[b]{1\linewidth}
  \centering
 \includegraphics[width=\textwidth]{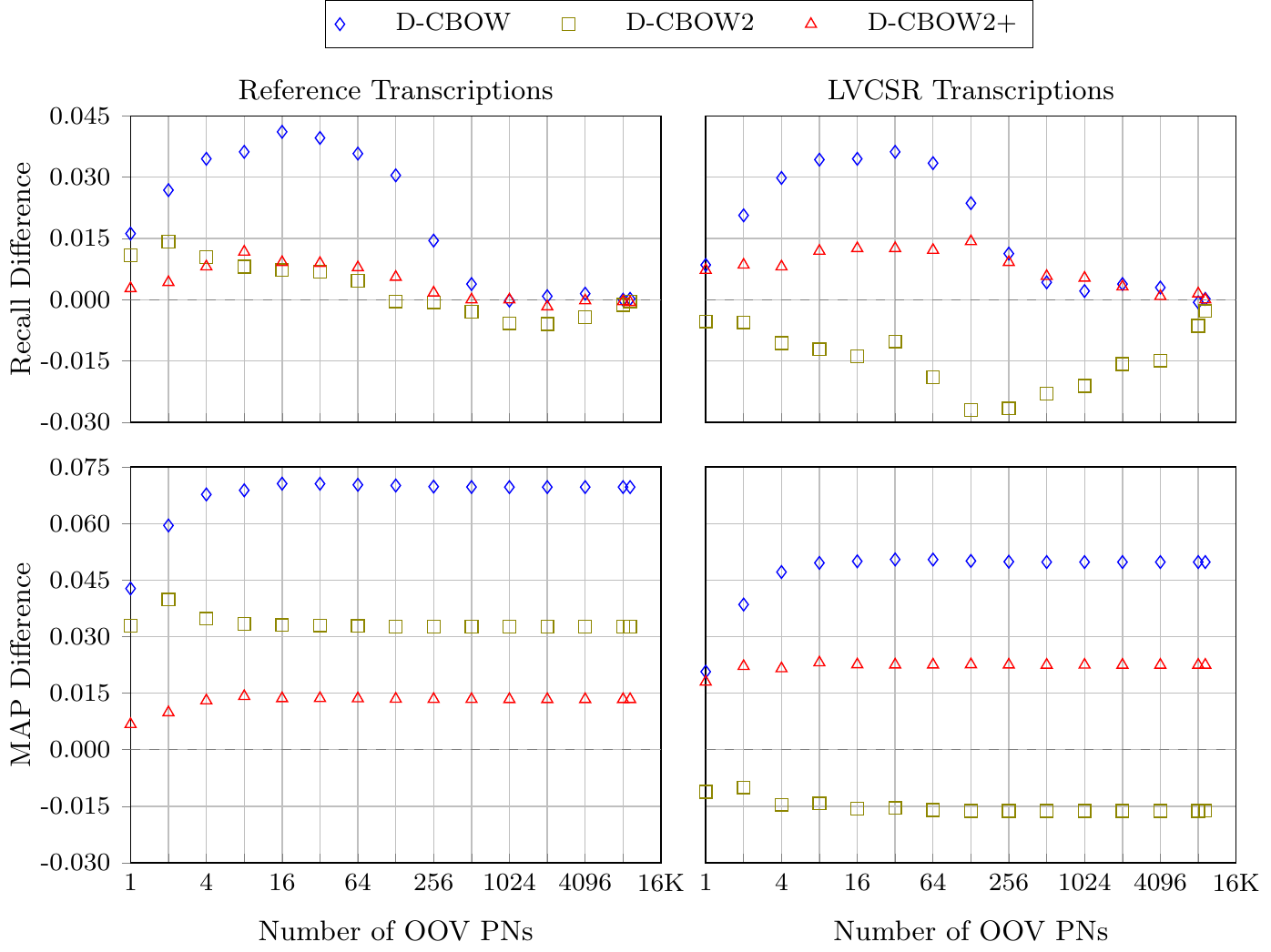}
\end{minipage}
\caption{Difference in Recall \& MAP performance of OOV PN retrieval between second and first training phase of D-CBOW, DCBOW2 and D-CBOW2+.}
\label{fig:diff}
\end{figure*}

\end{appendices}

\end{document}